\documentclass[11pt,a4paper]{article}
\usepackage{arabtex}
\usepackage{utf8}
\setcode{utf8}

\usepackage[utf8]{inputenc}
\usepackage[hyperref]{acl2021}
\usepackage[arabic, english]{babel}
\usepackage{times}
\usepackage{latexsym}

\usepackage{colortbl}
\usepackage{color}
\usepackage{array}
\usepackage{multirow}

\usepackage{microtype}
\usepackage[OT2,T1]{fontenc}
\usepackage{tipa}
\usepackage{sidecap}
\usepackage{booktabs}
\usepackage{pgfplots}
\pgfplotsset{compat=1.6}
\usepackage{subcaption}
\usepackage{caption}
\usepackage{adjustbox}

\defcitealias{b-etal-2010-resource}{Sowmya et al., 2010}

\newcommand{\Sref}[1]{\S\ref{#1}}

\newcommand{\Fref}[1]{Figure~\ref{#1}}

\newcommand{\Tref}[1]{Table~\ref{#1}}

\colorlet{TransparentRed}{white!70!red}
\colorlet{TransparentYellow}{white!50!yellow}
\newcommand{\errorbox}[1]{\setlength{\fboxsep}{1pt}\colorbox{TransparentRed}{#1}}
\newcommand{\translitbox}[1]{\setlength{\fboxsep}{1pt}\colorbox{TransparentYellow}{#1}}
\newcommand{\gap}{\errorbox{\vphantom{0}}}
\newcommand\textcyr[1]{{\fontencoding{OT2}\fontfamily{wncyr}\selectfont #1}}
\newcommand\sci[1]{{\small[{#1}]}}
\newcommand\parnospace[1]{\noindent\textbf{#1}}

\aclfinalcopy

\title{Comparative Error Analysis in Neural and Finite-state Models for Unsupervised Character-level Transduction}

\author{Maria Ryskina$^1$\quad Eduard Hovy$^1$\quad Taylor Berg-Kirkpatrick$^2$\quad Matthew R. Gormley$^3$\\
         $^1$Language Technologies Institute, Carnegie Mellon University \\
         $^2$Computer Science and Engineering, University of California, San Diego \\
         $^3$Machine Learning Department, Carnegie Mellon University \\ 
         \texttt{mryskina@cs.cmu.edu} \quad \texttt{hovy@cmu.edu}\\ 
         \texttt{tberg@eng.ucsd.edu} \quad \texttt{mgormley@cs.cmu.edu}
         } 

\date{}

\begin{document}
\maketitle
\begin{abstract}

Traditionally, character-level transduction problems have been solved with finite-state models designed to encode structural and linguistic knowledge of the underlying process, whereas recent approaches rely on the power and flexibility of sequence-to-sequence models with attention. Focusing on the less explored unsupervised learning scenario, we compare the two model classes side by side and find that they tend to make different types of errors even when achieving comparable performance. We analyze the distributions of different error classes using two unsupervised tasks as testbeds: converting informally romanized text into the native script of its language (for Russian, Arabic, and Kannada) and translating between a pair of closely related languages (Serbian and Bosnian). Finally, we investigate how combining finite-state and sequence-to-sequence models at decoding time affects the output quantitatively and qualitatively.\footnote{Code will be published at \url{https://github.com/ryskina/error-analysis-sigmorphon2021}} 

\end{abstract}

\section{Introduction and prior work}
\label{sec:intro}
Many natural language sequence transduction tasks, such as transliteration or grapheme-to-phoneme conversion, call for a character-level parameterization that reflects the linguistic knowledge of the underlying generative process. Character-level transduction approaches have even been shown to perform well for tasks that are not entirely character-level in nature, such as translating between related languages~\cite{pourdamghani-knight-2017-deciphering}.

\begin{figure}[t]
    \centering
    \includegraphics[width=\columnwidth]{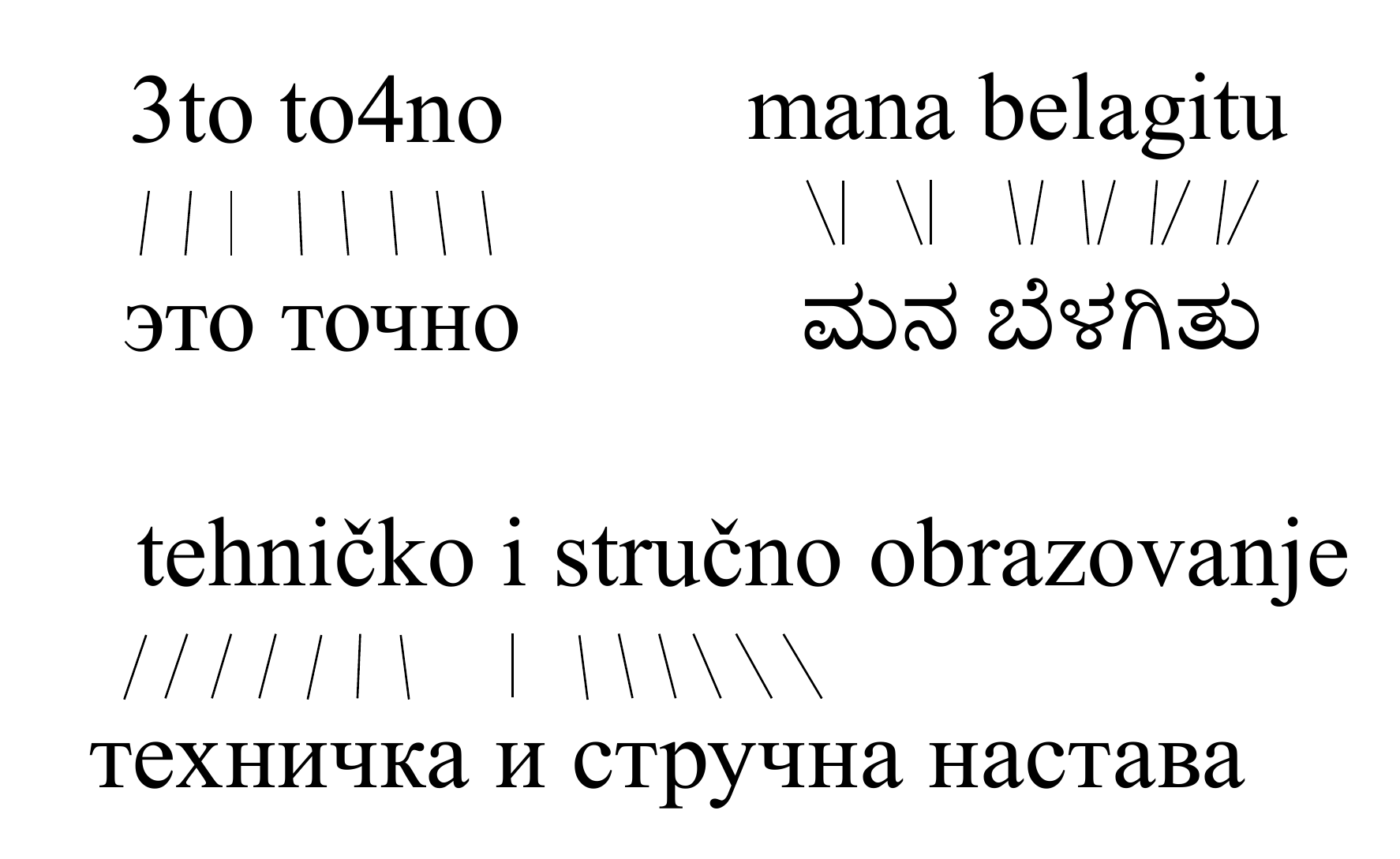}
    \caption{Parallel examples from our test sets for two character-level transduction tasks: converting informally romanized text to its original script (top; examples in Russian and Kannada) and translating between closely related languages (bottom; Bosnian--Serbian). Informal romanization is idiosyncratic and relies on both visual (\textcyr{ch} $\rightarrow$ 4) and phonetic (\textcyr{t} $\rightarrow$ t) character similarity, while translation is more standardized but not fully character-level due to grammatical and lexical differences (`\textcyr{nastava}' $\rightarrow$ `obrazovanje') between the languages. The lines show character alignment between the source and target side where possible.}
    \label{fig:tasks}
\end{figure}

Weighted finite-state transducers (WFSTs) have traditionally been used for such character-level tasks~\cite{knight-graehl-1998-machine, knight-etal-2006-unsupervised}. Their structured formalization
makes it easier to encode additional constraints, imposed either by the underlying linguistic process (e.g.\ monotonic character alignment) or by the probabilistic generative model (Markov assumption; \citealp{eisner-2002-parameter}).
Their interpretability also facilitates the introduction of useful inductive bias, which is crucial for unsupervised training~\cite{ravi-knight-2009-learning, ryskina-etal-2020-phonetic}.

Unsupervised neural sequence-to-sequence (seq2seq) architectures have also shown impressive performance on tasks like machine translation~\cite{lample2018unsupervised} and style transfer~\cite{yang2018unsupervised, he2020a}. These models are substantially more powerful than WFSTs, and they successfully learn the underlying patterns from monolingual data without
any explicit information about the underlying generative process.

As the strengths of the two model classes differ, so do their weaknesses: the WFSTs and the seq2seq models are prone to different kinds of errors. On a higher level, it is explained by the structure--power trade-off: while the seq2seq models are better at recovering long-range dependencies and their outputs look less noisy, they also tend to insert and delete words arbitrarily because their alignments are unconstrained.
We attribute the errors to the following aspects of the trade-off: 
\\[0.1in]
\parnospace{Language modeling capacity:} 
the statistical character-level n-gram language models (LMs) utilized by finite-state approaches are much weaker than the RNN language models with unlimited left context. While a word-level LM can improve the performance of a WFST, it would also restrict the model's ability to handle out-of-vocabulary words.
\\[0.1in]
\parnospace{Controllability of learning:} 
more structured models allow us to ensure that the model does not attempt to learn patterns orthogonal to the underlying process. For example, domain imbalance between the monolingual corpora can cause the seq2seq models to exhibit unwanted style transfer effects like inserting frequent target side words arbitrarily.
\\[0.1in]
\parnospace{Search procedure:} 
WFSTs make it easy to perform exact maximum likelihood decoding via shortest-distance algorithm~\cite{mohri2009weighted}. For the neural models trained using conventional methods, decoding strategies that optimize for the output likelihood (e.g.\ beam search with a large beam size) have been shown to be susceptible to favoring empty outputs~\cite{stahlberg-byrne-2019-nmt} and generating repetitions~\cite{holtzman2020the}.

\medskip
Prior work on leveraging the strength of the two approaches proposes complex joint parameterizations, such as neural weighting of WFST arcs or paths~\cite{rastogi-etal-2016-weighting, lin-etal-2019-neural} or encoding alignment constraints into the attention layer of seq2seq models~\cite{aharoni-goldberg-2017-morphological, wu-etal-2018-hard, wu-cotterell-2019-exact, makarov-etal-2017-align}. 
We study whether performance can be improved with simpler decoding-time model combinations, reranking and product of experts, which have been used effectively for other model classes~\cite{charniak-johnson-2005-coarse, hieber-riezler-2015-bag}, evaluating on two unsupervised tasks: decipherment of informal romanization~\cite{ryskina-etal-2020-phonetic} and related language translation~\cite{pourdamghani-knight-2017-deciphering}.

While there has been much error analysis for the WFST and seq2seq approaches separately, it largely focuses on the more common supervised case. We perform detailed side-by-side error analysis to draw high-level comparisons between finite-state and seq2seq models and investigate if the intuitions from prior work would transfer to the unsupervised transduction scenario.

\section{Tasks}

We compare the errors made by the finite-state and the seq2seq approaches by analyzing their performance on two unsupervised character-level transduction tasks: translating between closely related languages written in different alphabets and converting informally romanized text into its native script. Both tasks are illustrated in \Fref{fig:tasks}.

\subsection{Informal romanization}
Informal romanization is an idiosyncratic transformation that
renders a non-Latin-script language in Latin alphabet, extensively used online by speakers of Arabic~\cite{darwish-2014-arabizi}, Russian~\cite{paulsen20149}, and many Indic languages~\citepalias{b-etal-2010-resource}. \Fref{fig:tasks} shows examples of romanized Russian (top left) and Kannada (top right) sentences along with their ``canonicalized'' representations in Cyrillic and Kannada scripts respectively. Unlike official romanization systems such as pinyin, this type of transliteration is not standardized: character substitution choices vary between users and are based on the specific user's perception of how similar characters in different scripts are. Although the substitutions are primarily phonetic (e.g.\ Russian \textcyr{n}~\textipa{/n/}~$\rightarrow$~n), i.e. based on the pronunciation of a specific character in or out of context, users might also rely on visual similarity between glyphs (e.g.\ Russian \textcyr{ch}~\textipa{/\texttoptiebar{tS\textsuperscript{j}}/}~$\rightarrow$~4), especially when the associated phoneme cannot be easily mapped to a Latin-script grapheme (e.g.\ Arabic \<ع>~\textipa{/Q/}~$\rightarrow$~3). To capture this  variation, we view the task of decoding informal romanization as a many-to-many character-level decipherment problem.

The difficulty of deciphering romanization also depends on the type of the writing system the language traditionally uses. In alphabetic scripts, where grapheme-to-phoneme correspondence is mostly one-to-one, there tends to be a one-to-one monotonic alignment between characters in the romanized and native script sequences (\Fref{fig:tasks}, top left). \textit{Abjads} and \textit{abugidas}, where graphemes correspond to consonants or consonant-vowel syllables, increasingly use many-to-one alignment in their romanization (\Fref{fig:tasks}, top right), which makes learning the latent alignments, and therefore decoding, more challenging. In this work, we experiment with three languages spanning over three major types of writing systems---Russian (alphabetic),  Arabic (abjad), and Kannada (abugida)---and compare how well-suited character-level models are for learning these varying alignment patterns.

\subsection{Related language translation}

As shown by~\citet{pourdamghani-knight-2017-deciphering} and \citet{hauer-etal-2014-solving}, character-level models can be used effectively to translate between languages that are closely enough related to have only small lexical and grammatical differences, such as Serbian and Bosnian~\cite{ljubesic-klubicka-2014}. We focus on this specific language pair and tie the languages to specific orthographies (Cyrillic for Serbian and Latin for Bosnian), approaching the task as an unsupervised orthography conversion problem. However, the transliteration framing of the translation problem is inherently limited since the task is not truly character-level in nature, as shown by the alignment lines in \Fref{fig:tasks} (bottom). Even the most accurate transliteration model will not be able to capture non-cognate word translations (Serbian `\textcyr{nastava}'~\sci{nastava, `education, teaching'} $\rightarrow$ Bosnian `obrazovanje' \sci{`education'}) and the resulting discrepancies in morphological inflection (Serbian \textcyr{-a} endings in adjectives agreeing with feminine `\textcyr{nastava}' map to Bosnian -o representing agreement with neuter `obrazovanje').

One major difference with the informal romanization task is the lack of the idiosyncratic orthography: the word spellings are now consistent across the data. However, since the character-level approach does not fully reflect the nature of the transformation, the model will still have to learn a many-to-many cipher with highly context-dependent character substitutions.

\begin{table*}[t]
    \centering
    \resizebox{0.85\textwidth}{!}{%
        \begin{tabular}{@{}lrrrrrrrr@{}}
            \toprule
            & \multicolumn{2}{c}{Train (source)} & \multicolumn{2}{c}{Train (target)} & \multicolumn{2}{c}{Validation} & \multicolumn{2}{c}{Test}\\
            & \multicolumn{1}{c}{Sent.} & \multicolumn{1}{c}{Char.} & \multicolumn{1}{c}{Sent.} & \multicolumn{1}{c}{Char.} &
            \multicolumn{1}{c}{Sent.} & \multicolumn{1}{c}{Char.} &
            \multicolumn{1}{c}{Sent.} & \multicolumn{1}{c}{Char.}\\
            \midrule
           Romanized Arabic & 5K & 104K & 49K & 935K & 301 & 8K & 1K & 20K\\
           Romanized Russian & 5K & 319K & 307K & 111M & 227 & 15K & 1K & 72K \\
           Romanized Kannada & 10K & 1M & 679K & 64M & 100 & 11K & 100 & 10K \\
           Serbian$\rightarrow$Bosnian & 160K & 9M & 136K & 9M & 16K & 923K & 100 & 9K \\
           Bosnian$\rightarrow$Serbian & 136K & 9M & 160K & 9M & 16K & 908K & 100 & 10K \\
        \bottomrule
        \end{tabular}
    }
    \caption{Dataset splits for each task and language. The source and target train data are monolingual, and the validation and test sentences are parallel. For the informal romanization task, the source and target sides correspond to the Latin and the original script respectively. For the translation task, the source and target sides correspond to source and target languages. The validation and test character statistics are reported for the source side.}
    \label{tab:splits}
\end{table*}

\section{Data}

\Tref{tab:splits} details the statistics of the splits used for all languages and tasks. Below we describe each dataset in detail, explaining the differences in data split sizes between languages. Additional preprocessing steps applied to all datasets are described in \Sref{sec:preprocessing}.\footnote{Links to download the corpora and other data sources discussed in this section can be found in Appendix \ref{app:download}.}

\subsection{Informal romanization}

\begin{figure}[!h]
    \small
    \centering
    \setlength\tabcolsep{2pt}
    \resizebox{0.68\columnwidth}{!}{
    \begin{tabular}{p{0.2\columnwidth}p{0.38\columnwidth}}
        \midrule
         \textbf{Source:} 
            & de el menu\textcolor{red}{:)}
         \\
         \textbf{Filtered:} 
            & de \textcolor{blue}{el menu}<...>
         \\
         \textbf{Target:} 
            & \textcolor{black}{<...>}\textcolor{blue}{\< ألمنه>} \<دي>
         \\
         \textbf{Gloss:} 
            & `This is the menu'
         \\
        \midrule
    \end{tabular}
    }
    \caption{A parallel example from the LDC BOLT Arabizi dataset, written in Latin script (source) and converted to Arabic (target) semi-manually. Some source-side segments (in \textcolor{red}{red}) are removed by annotators; we use the version without such segments (filtered) for our task. The annotators also standardize spacing on the target side,  which results in difference with the source (in \textcolor{blue}{blue}).
    }
\label{fig:arabic-example}
\end{figure}

\paragraph{Arabic} We use the LDC BOLT Phase 2 corpus~\cite{bies-etal-2014-transliteration, song-etal-2014-collecting} for training and testing the Arabic transliteration models (\Fref{fig:arabic-example}). The corpus consists of short SMS and chat in Egyptian Arabic represented using Latin script (\emph{Arabizi}). The corpus is fully parallel: each message is automatically converted into the standardized dialectal Arabic orthography (CODA; \citealp{habash-etal-2012-conventional}) and then manually corrected by human annotators. We split and preprocess the data according to \citet{ryskina-etal-2020-phonetic}, discarding the target (native script) and source (romanized) parallel sentences to create the  source and target monolingual training splits respectively.

\definecolor{light-gray}{gray}{0.94}
\begin{figure}[!ht]
    \setlength\tabcolsep{2pt}
    \small
    \centering
    \resizebox{\columnwidth}{!}{
    \begin{tabular}{p{0.2\columnwidth}p{0.7\columnwidth}}
       \midrule
        \rowcolor{light-gray} \multicolumn{2}{c}{Annotated}\\
         \textbf{Source:} 
            & proishodit s prirodoy \textcolor{blue}{4to to} \textcolor{red}{very very bad}
         \\
         \textbf{Filtered:} 
            & proishodit s prirodoy \textcolor{blue}{4to to} <...>\\
         \textbf{Target:} 
            & \textcyr{proiskhodit s prirodo\u{i} \textcolor{blue}{chto-to}} <...>
         \\
         \textbf{Gloss:} 
            & `Something very very bad is happening to the environment' 
         \\
         \midrule
         \rowcolor{light-gray} \multicolumn{2}{c}{Monolingual}\\
         \textbf{Source:} & ---
          \\
         \textbf{Target:} 
            & \textcyr{e1to videoroliki so sp2ezda partii``Edinaya Rossiya''}
         \\
         \textbf{Gloss:} 
            & `These are the videos from the ``United Russia'' party congress'
         \\
        \midrule
    \end{tabular}
    }
    \caption{\textbf{Top:} A parallel example from the romanized Russian dataset. We use the filtered version of the romanized (source) sequences, removing the segments the annotators were unable to convert to Cyrillic, e.g.\ code-switched phrases (in \textcolor{red}{red}). The annotators also standardize minor spelling variation such as hyphenation (in \textcolor{blue}{blue}). 
    \textbf{Bottom:} a monolingual Cyrillic example from the \texttt{vk.com} portion of the Taiga corpus, which mostly consists of comments in political discussion groups.
    }
\label{fig:russian-example}
\end{figure}

\paragraph{Russian} We use the romanized Russian dataset collected by \citet{ryskina-etal-2020-phonetic}, augmented with the monolingual Cyrillic data from the Taiga corpus of~\citet{shavrina2017methodology} (\Fref{fig:russian-example}). The romanized data is split into training, validation, and test portions, and all validation and test sentences are converted to Cyrillic by native speaker annotators. Both the romanized and the native-script sequences are collected from public posts and comments on a Russian social network \texttt{vk.com}, and they are on average 3 times longer than the messages in the Arabic dataset (\Tref{tab:splits}). However, although both sides were scraped from the same online platform, the relevant Taiga data is collected primarily from political discussion groups, so there is still a substantial domain mismatch between the source and target sides of the data.

\begin{figure}[!ht]
    \small
    \centering
    \setlength\tabcolsep{2pt}
    \resizebox{\columnwidth}{!}{
    \begin{tabular}{p{0.2\columnwidth}p{0.7\columnwidth}}
        \midrule
         \textbf{Target:}
            & \raisebox{-.3\height}{\includegraphics[height=0.5cm]{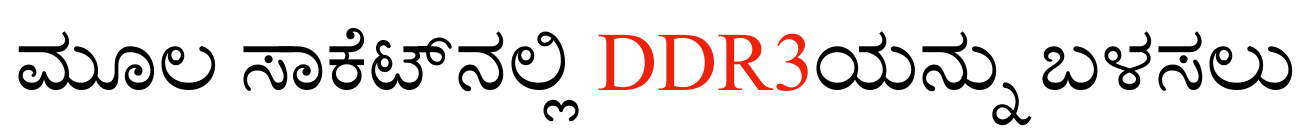}}
         \\
         \textbf{Source:} 
            & moola saaketnalli \textcolor{red}{ddr3}yannu balasalu 
         \\
         \textbf{Gloss:} 
            & `to use DDR3 in the source circuit'
         \\
         \midrule
    \end{tabular}
   }
   \caption{A parallel example from the Kannada portion of the Dakshina dataset. The Kannada script data (target) is scraped from Wikipedia and manually converted to Latin (source) by human annotators. Foreign target-side characters (in \textcolor{red}{red}) get preserved in the annotation but our preprocessing replaces them with UNK on the target side. \label{fig:kannada-example}}
\end{figure}

\paragraph{Kannada} Our Kannada data (\Fref{fig:kannada-example}) is taken from the Dakshina dataset~\cite{roark-etal-2020-dakshina}, a large collection of native-script text from Wikipedia for 12 South Asian languages. Unlike the Russian and Arabic data, the romanized portion of Dakshina is not scraped directly from the users' online communication, but instead elicited from native speakers given the native-script sequences. Because of this, all romanized sentences in the data are parallel: we allocate most of them to the source side training data, discarding their original script counterparts, and split the remaining annotated ones between validation and test.

\begin{figure}[!t]
    \centering
    \small
    \setlength\tabcolsep{2pt}
    \resizebox{\columnwidth}{!}{
    \begin{tabular}{p{0.2\columnwidth}p{0.7\columnwidth}}
        \midrule
         \textbf{Serbian:} 
            & \textcyr{svako ima pravo na zhivot, slobodu i \textcolor{blue}{bezbednost lichnosti}.}
         \\
         \textbf{Bosnian:} 
            & svako ima pravo na \v{z}ivot, slobodu i \textcolor{blue}{osobnu sigurnost}.
         \\
         \textbf{Gloss:} 
            & `Everyone has the right to life, liberty and security of person.'
         \\
         \midrule
    \end{tabular}
    }
    \caption{A parallel example from the Serbian--Cyrillic and Bosnian--Latin UDHR. The sequences are not entirely parallel on character level due to paraphrases and non-cognate translations (in \textcolor{blue}{blue}).}
\label{fig:srbos-example}
\end{figure}

\subsection{Related language translation}
Following prior work \cite{pourdamghani-knight-2017-deciphering, yang2018unsupervised, he2020a}, we train our unsupervised models on the monolingual data from the Leipzig corpora \cite{goldhahn-etal-2012-building}. We reuse the non-parallel training and synthetic parallel validation splits of \citet{yang2018unsupervised}, who generated their parallel data using the Google Translation API. Rather than using their synthetic test set, we opt to test on natural parallel data from the Universal Declaration of Human Rights (UDHR), following \citet{pourdamghani-knight-2017-deciphering}.

We manually sentence-align the Serbian--Cyrillic and Bosnian--Latin declaration texts and follow the preprocessing guidelines of~\citet{pourdamghani-knight-2017-deciphering}. Although we strive to approximate the training and evaluation setup of their work for fair comparison, there are some discrepancies: for example, our manual alignment of UDHR yields 100 sentence pairs compared to 104 of~\citet{pourdamghani-knight-2017-deciphering}.
We use the data to train the translation models in both directions, 
simply switching the source and target sides from Serbian to Bosnian and vice versa.

\subsection{Inductive bias}
\label{sec:priors}

As discussed in \Sref{sec:intro}, the WFST models are less powerful than the seq2seq models; however, they are also more  structured, which we can use to introduce inductive bias to aid unsupervised training. Following \citet{ryskina-etal-2020-phonetic}, we introduce informative priors on character substitution operations (for a description of the WFST parameterization, see \Sref{sec:base-models}). The priors reflect the visual and phonetic similarity between characters in different alphabets and are sourced from human-curated resources built with the same concepts of similarity in mind. For all tasks and languages, we collect phonetically similar character pairs from the phonetic keyboard layouts (or, in case of the translation task, from the default Serbian keyboard layout, which is phonetic in nature due to the dual orthography standard of the language). We also add some visually similar character pairs by automatically pairing all symbols that occur in both source and target alphabets (same Unicode codepoints). For Russian, which exhibits a greater degree of visual similarity than Arabic or Kannada, we also make use of the Unicode confusables list (different Unicode codepoints but same or similar glyphs).\footnote{Links to the keyboard layouts and the confusables list can be found in Appendix \ref{app:download}.}

It should be noted that these automatically generated informative priors also contain noise: keyboard layouts have spurious mappings because each symbol must be assigned to exactly one key in the QWERTY layout, and Unicode-constrained visual mappings might prevent the model from learning correspondences between punctuation symbols (e.g. Arabic question mark \<؟>~$\rightarrow$~?).

\subsection{Preprocessing}
\label{sec:preprocessing}
We lowercase and segment all sequences into characters as defined by Unicode codepoints, so diacritics and non-printing characters like ZWJ are also treated as separate vocabulary items. To filter out foreign or archaic characters and rare diacritics, we restrict the alphabets to characters that cover 99\% of the monolingual training data. After that, we add any standard alphabetical characters and numerals that have been filtered out back into the source and target alphabets. All remaining filtered characters are replaced with a special UNK symbol in all splits except for the target-side test.

\begin{table*}[!t]
\centering
\resizebox{0.9\textwidth}{!}{%
    \begin{tabular}{@{}lccccccccc@{}}
         \toprule
         & \multicolumn{3}{c}{Arabic} & \multicolumn{3}{c}{Russian} & \multicolumn{3}{c}{Kannada} \\
         & CER & WER & BLEU & CER & WER & BLEU & CER & WER & BLEU\\
         \midrule
         WFST & .405 & .86 & 2.3 & .202 & .58 & 14.8 & .359 & .\textbf{71} & 12.5 \\
         Seq2Seq & .571 & .85 & 4.0 & .229 & .\textbf{38} & \textbf{48.3} & .559 & .79 & 11.3\\ 
         \midrule
         Reranked WFST & .\textbf{398} & .85 & 2.8 & .195 & .57 & 16.1 & .\textbf{358} & .\textbf{71} & 12.5\\
         Reranked Seq2Seq & .538 & .\textbf{82} & \textbf{4.6} & .216 & .39 & 45.6 & .545 & .78 & \textbf{12.6}\\
         Product of experts & .470 & .88 & 2.5 & .\textbf{178} & .50 & 22.9 & .543 & .93 & \hphantom{0}7.0\\
         \bottomrule
    \end{tabular}
    }
\caption{Character and word error rates (lower is better) and BLEU scores (higher is better) for the romanization decipherment task. \textbf{Bold} indicates best per column. Model combinations mostly interpolate between the base models' scores, although reranking yields minor improvements in character-level and word-level metrics for the WFST and seq2seq respectively. \textbf{Note:} base model results are not intended as a direct comparison between the WFST and seq2seq, since they are trained on different amounts of data.}
\label{tab:roman}
\end{table*}

{
\begin{table*}[!t]
\centering
\resizebox{0.8\textwidth}{!}{%
    \begin{tabular}{@{}lcccccc@{}}
         \toprule
         & \multicolumn{3}{c}{srp$\rightarrow$bos} & \multicolumn{3}{c}{bos$\rightarrow$srp}\\
         & CER & WER & BLEU & CER & WER & BLEU \\
         \midrule
         WFST &  .\textbf{314} & .50 & 25.3 & .319 & .52 & 25.5 \\
         Seq2Seq & .375 & .49 & 34.5 & .395 & .49 & 36.3\\ 
         \midrule
         Reranked WFST & .\textbf{314} & .49 & 26.3 & .\textbf{317} & .50 & 28.1 \\
         Reranked Seq2Seq & .376 & .\textbf{48} & 35.1 & .401 & .\textbf{47} & 37.0 \\
         Product of experts & .329 & .54 & 24.4 & .352 & .66 & 20.6 \\
         \midrule
         \cite{pourdamghani-knight-2017-deciphering} & --- & --- & \textbf{42.3} & --- & --- & \textbf{39.2} \\
          \cite{he2020a} & .657 & .81 & \hphantom{0}5.6 & .693 & .83 & \hphantom{0}4.7 \\
         \bottomrule
    \end{tabular}
}
\caption{Character and word error rates (lower is better) and BLEU scores (higher is better) for the related language translation task. \textbf{Bold} indicates best per column. The WFST and the seq2seq have comparable CER and WER despite the WFST being trained on up to 160x less source-side data (\Sref{sec:base-models}). While none of our models achieve the scores reported by \citet{pourdamghani-knight-2017-deciphering}, they all substantially outperform the subword-level model of~\citet{he2020a}. \textbf{Note:} base model results are not intended as a direct comparison between the WFST and seq2seq, since they are trained on different amounts of data.}
\label{tab:trans}
\end{table*}
}

\section{Methods}

We perform our analysis using the finite-state and seq2seq models from prior work and experiment with two joint decoding strategies, reranking and product of experts. Implementation details and hyperparameters are described in Appendix~\ref{app:implementation}.

\subsection{Base models}
\label{sec:base-models}

Our finite-state model is the WFST cascade introduced by~\citet{ryskina-etal-2020-phonetic}. The model is composed of a character-level n-gram language model and a script conversion transducer (emission model), which supports one-to-one character substitutions, insertions, and deletions. Character operation weights in the emission model are parameterized with multinomial distributions, and similar character mappings (\Sref{sec:priors}) are used to create Dirichlet priors on the emission parameters. To avoid marginalizing over sequences of infinite length, a fixed limit is set on the delay of any path (the difference between the cumulative number of insertions and deletions at any timestep). \citet{ryskina-etal-2020-phonetic} train the WFST using stochastic stepwise EM \cite{liang-klein-2009-online}, marginalizing over all possible target sequences and their alignments with the given source sequence. To speed up training, we modify their training procedure towards `hard EM': given a source sequence, we predict the most probable target sequence under the model, marginalize over alignments and then update the parameters. Although the unsupervised WFST training is still slow, the stepwise training procedure is designed to converge using fewer data points, so we choose to train the WFST model only on the 1,000 shortest source-side training sequences (500 for Kannada). 

Our default seq2seq model is the unsupervised neural machine translation (UNMT) model of~\citet{lample2018unsupervised, lample2019multipleattribute} in the parameterization of~\citet{he2020a}. 
The model consists of an LSTM~\cite{hochreiter1997long}
encoder and decoder with attention, trained to map sentences from each domain into a shared latent space. Using a combined objective, the UNMT model is trained to denoise, translate in both directions, and discriminate between the latent representation of sequences from different domains. Since the sufficient amount of balanced data is crucial for the UNMT performance, we train the seq2seq model on all available data on both source and target sides. Additionally, the seq2seq model decides on early stopping by evaluating on a small parallel validation set, which our WFST model does not have access to.

The WFST model treats the target and source training data differently, using the former to train the language model and the latter for learning the emission parameters, while the UNMT model is trained to translate in both directions simultaneously. Therefore, we reuse the same seq2seq model for both directions of the translation task, but train a separate finite-state model for each direction.

\begin{table*}[t]
    \centering
    \small
    \setlength\tabcolsep{2pt}
    \begin{tabular}{p{1in}>{\raggedright\arraybackslash}p{5in}}
        \toprule
        \rowcolor{light-gray} Input & 
            \textcyr{svako ima pravo da slobodno uchestvuje u kulturnom zhivotu zajednice, da uzhiva u umetnosti i da uchestvuje u nauchnom napretku i u dobrobiti koja otuda proistiche.}
        \\
        Ground truth & 
            svako ima pravo da slobodno sudjeluje u kulturnom \v{z}ivotu zajednice, da u\v{z}iva u umjetnosti i da u\v{c}estvuje u znanstvenom napretku i u njegovim koristima.
        \\
        \midrule
        \rowcolor{light-gray} WFST &
            svako ima pravo da slobodno \translitbox{u\v{c}estvuje} u kulturnom \v{z}ivotu \errorbox{s}jednice , da u\v{z}iva u \gap{}m\gap{}etnosti i da u\v{c}estvuje u \translitbox{nau\v{c}nom} napretku i u \translitbox{dobrobiti koja otuda pr\gap{}isti\v{c}e} .
        \\
        Reranked WFST &
            svako ima pravo da slobodno \translitbox{u\v{c}estvuje} u kulturnom \v{z}ivotu \errorbox{s}jednice , da u\v ziva u \errorbox{$\vphantom{0}$}m\errorbox{$\vphantom{0}$}etnosti i da u\v{c}estvuje u \translitbox{nau\v{c}nom} napretku i u \translitbox{dobrobiti koja otuda pr\errorbox{$\vphantom{0}$}isti\v{c}e} .
        \\
        \rowcolor{light-gray} Seq2Seq &
            svako ima pravo da slobodno \translitbox{u\v{c}estvuje} u kulturnom \v{z}ivotu zajednice , da \errorbox{$\phantom{000000000000000000}$} u\v{c}estvuje u \translitbox{nau\v{c}nom} napretku i u \translitbox{dobrobiti koja otuda proisti\v{c}e} .
        \\
        Reranked Seq2Seq &
            svako ima pravo da slobodno \translitbox{u\v{c}estvuje} u kulturnom \v{z}ivotu zajednice , da u\v ziva u umjetnosti i da u\v{c}estvuje u \translitbox{nau\v{c}nom} napretku i u \translitbox{dobrobiti koja otuda proisti\v{c}e}
        \\
        \rowcolor{light-gray} Product of experts &
            svako ima pravo da slobodno \translitbox{u\v{c}estvuje} u kulturnom \errorbox{za} u \errorbox{s}ajednice , da \gap{}\v{z}iva u umjetnosti i da u\v{c}estvuje u \translitbox{nau\v{c}nom} napretku i u \translitbox{dobro\errorbox{j }i koja otuda proisti\gap{}}
        \\
        Subword~Seq2Seq & 
            s\errorbox{ami} ima pravo da slobodno u\errorbox{ti\v{c}e na srpskom nivou vlasti} da \errorbox{razgovaraju} u \errorbox{bosne} i da \errorbox{djeluje} u \errorbox{me\dj unarodnom turizmu} i \errorbox{na buducnosti koja mu\v{z}a decisno} .
        \\
        \bottomrule
    \end{tabular}
    \caption{Different model outputs for a srp$\rightarrow$bos translation example. Prediction errors are highlighted in \errorbox{red}. Correctly transliterated segments that do not match the ground truth (e.g. due to paraphrasing) are shown in \translitbox{yellow}. Here the WFST errors are substitutions or deletions of individual characters, while the seq2seq drops entire words from the input (\Sref{sec:results} \#4). The latter problem is solved by reranking with a WFST for this example. The seq2seq model with subword tokenization~\cite{he2020a} produces mostly hallucinated output (\Sref{sec:results} \#2).  Example outputs for all other datasets can be found in the Appendix.}
    \label{tab:outputs}
\end{table*}

\subsection{Model combinations}
The simplest way to combine two independently trained models is reranking: using one model to produce a list of candidates and rescoring them according to another model. To generate candidates with a WFST, we apply the $n$--shortest paths algorithm~\cite{mohri2002efficient}. It should be noted that the $n$--best list might contain duplicates since each path represents a specific source--target character alignment. The length constraints encoded in the WFST also restrict its capacity as a reranker: beam search in the UNMT model may produce hypotheses too short or long to have a non-zero probability under the WFST.

Our second approach is a product-of-experts-style joint decoding strategy~\cite{hinton2002poe}: we perform beam search on the WFST lattice, reweighting the arcs with the output distribution of the seq2seq decoder at the corresponding timestep. 
For each partial hypothesis, we keep track of the WFST state $s$ and the partial input and output sequences $x_{1:k}$ and $y_{1:t}$.\footnote{Due to insertions and deletions in the emission model, $k$ and $t$ might differ; epsilon symbols are not counted.} When traversing an arc with input label $i\in \{x_{k+1}, \epsilon\}$ and output label $o$, we multiply the arc weight by the probability of the neural model outputting $o$ as the next character: $p_{\mathrm{seq2seq}}(y_{t+1}=o|x, y_{1:t})$. Transitions with $o=\epsilon$ (i.e.\ deletions) are not rescored by the seq2seq. We group hypotheses by their consumed input length $k$ and select $n$ best extensions at each timestep. 

\begin{SCfigure*}[][t]
    \centering
    \includegraphics[width=0.65\textwidth]{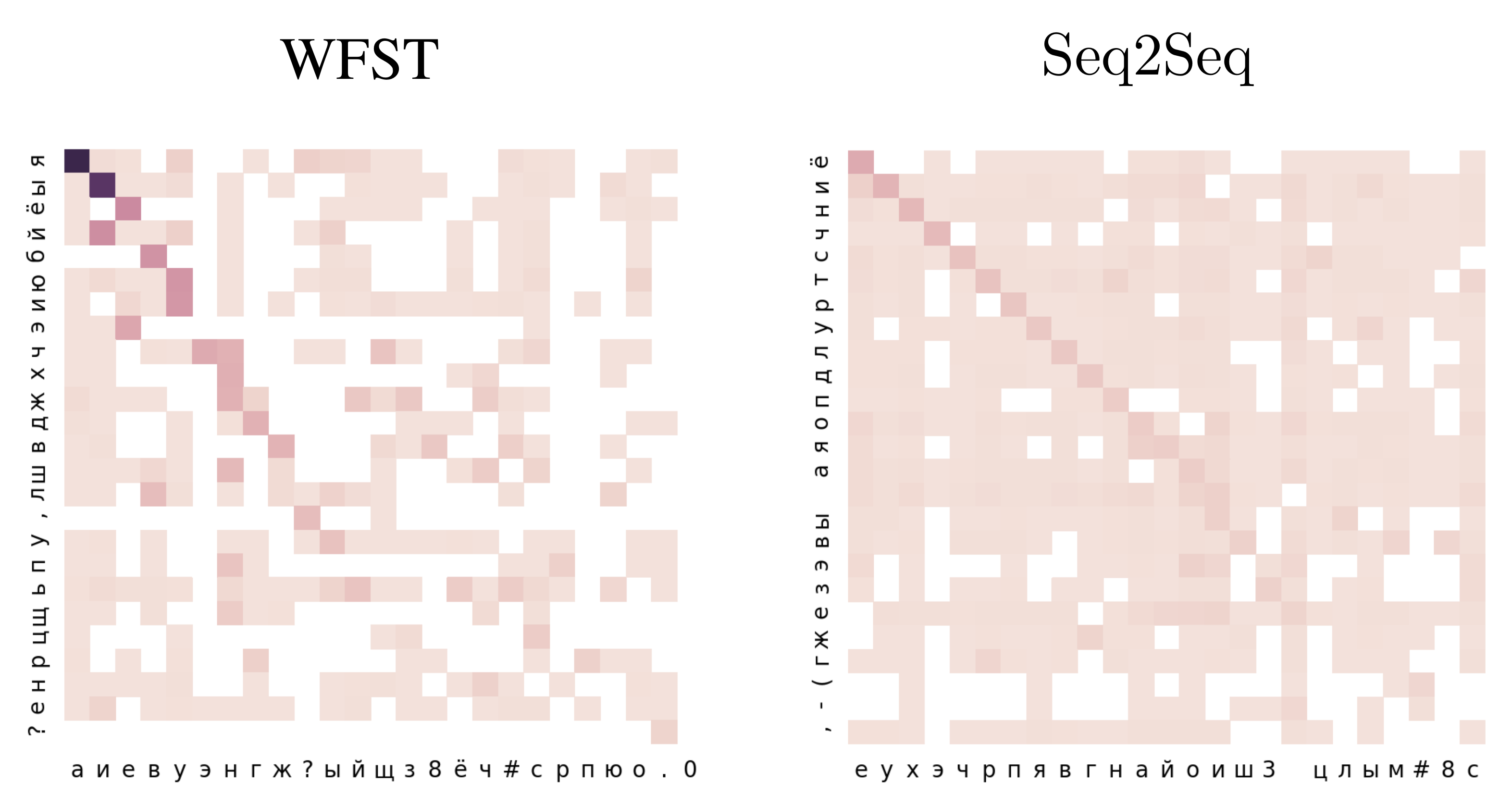}
    \caption{Highest-density submatrices of the two base models' character confusion matrices, computed in the Russian romanization task. White cells represent zero elements. The WFST confusion matrix (left) is noticeably sparser than the seq2seq one (right), indicating more repetitive errors. \# symbol stands for UNK.}
    \label{fig:heatmap}
\end{SCfigure*}

\subsection{Additional baselines}
For the translation task, we also compare to prior unsupervised approaches of different granularity: the deep generative style transfer model of~\citet{he2020a} and the character- and word-level WFST decipherment model of~\citet{pourdamghani-knight-2017-deciphering}. The former is trained on the same training set tokenized into subword units~\cite{sennrich-etal-2016-neural}, and we evaluate it on our UDHR test set for fair comparison. While the train and test data of~\citet{pourdamghani-knight-2017-deciphering} also use the same respective sources, we cannot account for tokenization differences that could affect the scores reported by the authors.

\section{Results and analysis}
\label{sec:results}

Tables~\ref{tab:roman} and \ref{tab:trans} present our evaluation of the two base models and three decoding-time model combinations on the romanization decipherment and related language translation tasks respectively. For each experiment, we report character error rate, word error rate, and BLEU (see Appendix~\ref{app:metrics}). The results for the base models support what we show later in this section: the seq2seq model is more likely to recover words correctly (higher BLEU, lower WER), while the WFST is more faithful on character level and avoids word-level substitution errors (lower CER). Example predictions can be found in~\Tref{tab:outputs} and in the Appendix.

Our further qualitative and quantitative findings are summarized in the following high-level takeaways:

\smallskip
\paragraph{\#1: Model combinations still suffer from search issues.}

We would expect the combined decoding to discourage all errors common under one model but not the other, improving the performance by leveraging the strengths of both model classes. However, as Tables~\ref{tab:roman} and \ref{tab:trans} show, they instead mostly interpolate between the scores of the two base models. In the reranking experiments, we find that this is often due to the same base model error (e.g.\ the seq2seq model hallucinating a word mid-sentence) repeating across all the hypotheses in the final beam. This suggests that successful reranking would require a much larger beam size or a diversity-promoting search mechanism. 

Interestingly, we observe that although adding a reranker on top of a decoder does improve performance slightly, the gain is only
in terms of the metrics that the base decoder is already strong at---character-level for reranked WFST and word-level for reranked seq2seq---at the expense of the other scores. Overall, none of our decoding strategies achieves best results across the board, and no model combination substantially outperforms both base models in any metric.

\smallskip
\paragraph{\#2: Character tokenization boosts performance of the neural model.}
In the past, UNMT-style models have been applied to various unsupervised sequence transduction problems. However, since these models were designed to operate on word or subword level, prior work assumes the same tokenization is necessary. We show that for the tasks allowing character-level framing, such models in fact respond extremely well to character input.

\Tref{tab:trans} compares the UNMT model trained on characters with the seq2seq style transfer model of~\citet{he2020a} trained on subword units. The original paper shows improvement over the UNMT baseline in the same setting, but simply switching to character-level tokenization without any other changes results in a 30 BLEU points gain for either direction. This suggests that the tokenization choice could act as an inductive bias for seq2seq models, and character-level framing could be useful even for tasks that are not truly character-level. 

This observation also aligns with the findings of the recent work on language modeling complexity~\cite{10.1162/tacl_a_00365, mielke-etal-2019-kind}. For many languages, including several Slavic ones related to the Serbian--Bosnian pair, a character-level language model yields lower surprisal than the one trained on BPE units, suggesting that the effect might also be explained by the character tokenization making the language easier to language-model.

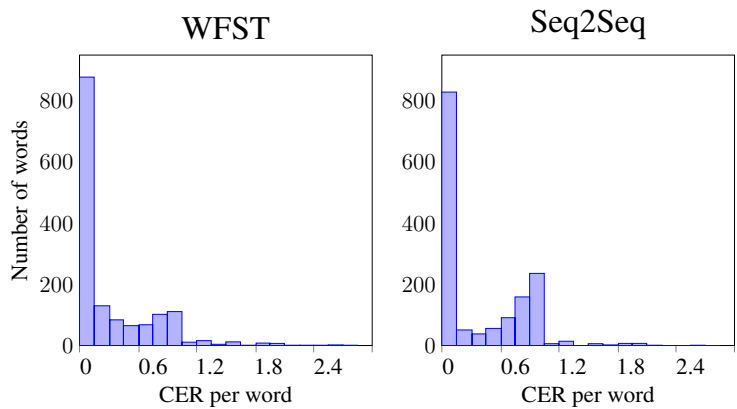
\begin{SCfigure*}[][t]
    \centering
        \subfloat{
            \scalebox{.6}{
                \begin{tikzpicture}
                \pgfplotsset{every tick label/.append style={font=\Large}
                }
                \begin{axis}[ybar interval,ymax=950, ymin=0,
                            xmin=0.07, xmax=2.92+0.15, grid=none,                          xtick={0.07,0.68,1.27,1.88,2.47,2.92+0.15},
                            xticklabels={0,0.6,1.2,1.8,2.4,x},
                            xticklabel style={text height=2ex, text width=0.5cm,anchor=north east,},
                            xtick pos=bottom,
                            ytick pos=left,
                            title={\huge WFST},
                            ylabel={\Large Number of words},
                            xlabel={\Large CER per word},
                            width=8cm, height=8cm,
                            ]
                \addplot coordinates {(0.07, 878.0) (0.22, 130.0) (0.38, 84.0) (0.52, 65.0) (0.68, 68.0) (0.82, 102.0) (0.97, 111.0) (1.12, 11.0) (1.27, 16.0) (1.42, 4.0) (1.57, 12.0) (1.72, 1.0) (1.88, 8.0) (2.02, 7.0) (2.17, 1.0) (2.33, 1.0) (2.47, 1.0) (2.62, 2.0) (2.77, 1.0) (2.92, 1.0)};
                \end{axis}
                \end{tikzpicture}
            }
        }
        \subfloat{
            \scalebox{.6}{
                \begin{tikzpicture}
                \pgfplotsset{every tick label/.append style={font=\Large}}
                \begin{axis}[ybar interval,ymax=950, ymin=0,
                            xmin=0.07, xmax=2.92+0.15, grid=none, xtick={0.07,0.68,1.27,1.88,2.47,2.92+0.15},
                            xticklabels={0,0.6,1.2,1.8,2.4,x},
                            xticklabel style={text height=2ex, text width=0.5cm,anchor=north east,},
                            xtick pos=bottom,
                            ytick pos=left,
                            title={\huge Seq2Seq},
                            width=8cm, height=8cm,
                            xlabel={\Large CER per word},
                            ]
                \addplot coordinates {(0.07, 829.0) (0.22, 51.0) (0.38, 38.0) (0.52, 56.0) (0.68, 91.0) (0.82, 159.0) (0.97, 236.0) (1.12, 6.0) (1.27, 14.0) (1.42, 0.0) (1.57, 6.0) (1.72, 2.0) (1.88, 7.0) (2.02, 7.0) (2.17, 1.0) (2.33, 0.0) (2.47, 0.0) (2.62, 1.0) (2.77, 0.0) (2.92, 2.0)};
                \end{axis}
                \end{tikzpicture}
            }
        }
    \caption{Character error rate per word for the WFST (left) and seq2seq (right) bos$\rightarrow$srp translation outputs. The predictions are segmented using Moses tokenizer~\cite{koehn-etal-2007-moses} and aligned to ground truth with word-level edit distance. The increased frequency of CER=1 for the seq2seq model as compared to the WFST indicates that it replaces entire words more often.}
    \label{fig:hist}
\end{SCfigure*}

\smallskip
\paragraph{\#3: WFST model makes more repetitive errors.}
Although two of our evaluation metrics, CER and WER, are based on edit distance, they do not distinguish between the different types of edits (substitutions, insertions and deletions). Breaking them down by the edit operation, we find that while both models favor substitutions on both word and character levels, insertions and deletions are more frequent under the neural model (43\% vs.\ 30\% of all edits on the Russian romanization task).
We also find that the character substitution choices of the neural model are more context-dependent: while the total counts of substitution errors for the two models are comparable, the WFST is more likely to repeat the same few substitutions per character type. This is illustrated by \Fref{fig:heatmap}, which visualizes the most populated submatrices of the confusion matrices for the same task as heatmaps. The WFST confusion matrix is noticeably more sparse, with the same few substitutions occurring much more frequently than others: for example, WFST often mistakes \textcyr{ya} for \textcyr{a} and rarely for other characters, while the neural model's substitutions of \textcyr{ya} are distributed closer to uniform. This suggests that the WFST errors might be easier to correct with rule-based postprocessing. Interestingly, we did not observe the same effect for the translation task, likely due to a more constrained nature of the orthography conversion.

\smallskip
\paragraph{\#4: Neural model is more sensitive to data distribution shifts.}
The language model aiming to replicate its training data distribution could cause the output to deviate from the input significantly. This could be an artifact of a domain shift, such as in Russian, where the LM training data came from a political discussion forum: the seq2seq model frequently predicts unrelated domain-specific proper names in place of very common Russian words, e.g.\ \textcyr{zhiznp1}~\sci{\v zizn, `life'}~$\rightarrow$~\textcyr{Zyuganov}~\sci{Zjuganov, `Zyuganov (politician's last name)'} or \textcyr{\symbol{11}to}~\sci{\`{e}to, `this'}~$\rightarrow$~\textcyr{Edinaya Rossiya}~\sci{Edinaja Rossija, `United Russia (political party)'}, presumably distracted by the shared first character in the romanized version. To quantify the effect of a mismatch between the train and test data distributions in this case, we inspect the most common word-level substitutions under each decoding strategy, looking at all substitution errors covered by the 1,000 most frequent substitution `types' (ground truth--prediction word pairs) under the respective decoder.
We find that 25\% of the seq2seq substitution errors fall into this category, as compared to merely 3\% for the WFST---notable given the relative proportion of in-vocabulary words in the models' outputs (89\% for UNMT vs.\ 65\% for WFST).

Comparing the error rate distribution across output words for the translation task also supports this observation. As can be seen from \Fref{fig:hist}, the seq2seq model is likely to either predict the word correctly (CER of 0) or entirely wrong (CER of 1), while the the WFST more often predicts the word partially correctly---examples in \Tref{tab:outputs} illustrate this as well. We also see this in the Kannada outputs: WFST typically gets all the consonants right but makes mistakes in the vowels, while the seq2seq tends to replace the entire word. 

\section{Conclusion}
We perform comparative error analysis in finite-state and seq2seq models and their combinations for two unsupervised character-level tasks, informal romanization decipherment and related language translation. 
We find that the two model types tend towards different errors: seq2seq models are more prone to word-level errors caused by distributional shifts while WFSTs produce more character-level noise despite the hard alignment constraints. 

Despite none of our simple decoding-time combinations substantially outperforming the base models, we believe that combining neural and finite-state models to harness their complementary advantages is a promising research direction. Such combinations might involve biasing seq2seq models towards WFST-like behavior via pretraining or directly encoding constraints such as hard alignment or monotonicity into their parameterization \cite{wu-etal-2018-hard, wu-cotterell-2019-exact}. 
Although recent work has shown that the Transformer can learn to perform character-level transduction without such biases in a supervised setting \cite{wu-etal-2021-applying}, exploiting the structured nature of the task could be crucial for making up for the lack of large parallel corpora in low-data and/or unsupervised scenarios. We hope that our analysis provides insight into leveraging the strengths of the two approaches for modeling character-level phenomena in the absence of parallel data.

\section*{Acknowledgments} The authors thank Badr Abdullah, Deepak Gopinath, Junxian He, Shruti Rijhwani, and Stas Kashepava for helpful discussion, and the anonymous reviewers for their valuable feedback.

\bibliographystyle{acl_natbib}
\bibliography{anthology,refs}

\begin{thebibliography}{46}
\expandafter\ifx\csname natexlab\endcsname\relax\def\natexlab#1{#1}\fi

\bibitem[{Aharoni and Goldberg(2017)}]{aharoni-goldberg-2017-morphological}
Roee Aharoni and Yoav Goldberg. 2017.
\newblock \href {https://doi.org/10.18653/v1/P17-1183} {Morphological
  inflection generation with hard monotonic attention}.
\newblock In \emph{Proceedings of the 55th Annual Meeting of the Association
  for Computational Linguistics (Volume 1: Long Papers)}, pages 2004--2015,
  Vancouver, Canada. Association for Computational Linguistics.

\bibitem[{Allauzen et~al.(2007)Allauzen, Riley, Schalkwyk, Skut, and
  Mohri}]{openfst}
Cyril Allauzen, Michael Riley, Johan Schalkwyk, Wojciech Skut, and Mehryar
  Mohri. 2007.
\newblock Open{F}st: A general and efficient weighted finite-state transducer
  library.
\newblock In \emph{Proceedings of the Ninth International Conference on
  Implementation and Application of Automata, (CIAA 2007)}, volume 4783 of
  \emph{Lecture Notes in Computer Science}, pages 11--23. Springer.
\newblock \url{http://www.openfst.org}.

\bibitem[{B. et~al.(2010)B., Choudhury, Bali, Dasgupta, and
  Basu}]{b-etal-2010-resource}
Sowmya~V. B., Monojit Choudhury, Kalika Bali, Tirthankar Dasgupta, and Anupam
  Basu. 2010.
\newblock \href
  {http://www.lrec-conf.org/proceedings/lrec2010/pdf/182_Paper.pdf} {Resource
  creation for training and testing of transliteration systems for {I}ndian
  languages}.
\newblock In \emph{Proceedings of the Seventh International Conference on
  Language Resources and Evaluation ({LREC}'10)}, Valletta, Malta. European
  Language Resources Association (ELRA).

\bibitem[{Bies et~al.(2014)Bies, Song, Maamouri, Grimes, Lee, Wright, Strassel,
  Habash, Eskander, and Rambow}]{bies-etal-2014-transliteration}
Ann Bies, Zhiyi Song, Mohamed Maamouri, Stephen Grimes, Haejoong Lee, Jonathan
  Wright, Stephanie Strassel, Nizar Habash, Ramy Eskander, and Owen Rambow.
  2014.
\newblock \href {https://doi.org/10.3115/v1/W14-3612} {Transliteration of
  {A}rabizi into {A}rabic orthography: Developing a parallel annotated
  {A}rabizi-{A}rabic script {SMS}/chat corpus}.
\newblock In \emph{Proceedings of the {EMNLP} 2014 Workshop on {A}rabic Natural
  Language Processing ({ANLP})}, pages 93--103, Doha, Qatar. Association for
  Computational Linguistics.

\bibitem[{Charniak and Johnson(2005)}]{charniak-johnson-2005-coarse}
Eugene Charniak and Mark Johnson. 2005.
\newblock \href {https://doi.org/10.3115/1219840.1219862} {Coarse-to-fine
  n-best parsing and {M}ax{E}nt discriminative reranking}.
\newblock In \emph{Proceedings of the 43rd Annual Meeting of the Association
  for Computational Linguistics ({ACL}{'}05)}, pages 173--180, Ann Arbor,
  Michigan. Association for Computational Linguistics.

\bibitem[{Darwish(2014)}]{darwish-2014-arabizi}
Kareem Darwish. 2014.
\newblock \href {https://doi.org/10.3115/v1/W14-3629} {{A}rabizi detection and
  conversion to {A}rabic}.
\newblock In \emph{Proceedings of the {EMNLP} 2014 Workshop on {A}rabic Natural
  Language Processing ({ANLP})}, pages 217--224, Doha, Qatar. Association for
  Computational Linguistics.

\bibitem[{Eisner(2002)}]{eisner-2002-parameter}
Jason Eisner. 2002.
\newblock \href {https://doi.org/10.3115/1073083.1073085} {Parameter estimation
  for probabilistic finite-state transducers}.
\newblock In \emph{Proceedings of the 40th Annual Meeting of the Association
  for Computational Linguistics}, pages 1--8, Philadelphia, Pennsylvania, USA.
  Association for Computational Linguistics.

\bibitem[{Goldhahn et~al.(2012)Goldhahn, Eckart, and
  Quasthoff}]{goldhahn-etal-2012-building}
Dirk Goldhahn, Thomas Eckart, and Uwe Quasthoff. 2012.
\newblock \href
  {http://www.lrec-conf.org/proceedings/lrec2012/pdf/327_Paper.pdf} {Building
  large monolingual dictionaries at the {L}eipzig corpora collection: From 100
  to 200 languages}.
\newblock In \emph{Proceedings of the Eighth International Conference on
  Language Resources and Evaluation ({LREC}'12)}, pages 759--765, Istanbul,
  Turkey. European Language Resources Association (ELRA).

\bibitem[{Gorman(2016)}]{gorman2016pynini}
Kyle Gorman. 2016.
\newblock \href {https://doi.org/10.18653/v1/W16-2409} {{P}ynini: A {P}ython
  library for weighted finite-state grammar compilation}.
\newblock In \emph{Proceedings of the {SIGFSM} Workshop on Statistical {NLP}
  and Weighted Automata}, pages 75--80, Berlin, Germany. Association for
  Computational Linguistics.

\bibitem[{Habash et~al.(2012)Habash, Diab, and
  Rambow}]{habash-etal-2012-conventional}
Nizar Habash, Mona Diab, and Owen Rambow. 2012.
\newblock \href
  {http://www.lrec-conf.org/proceedings/lrec2012/pdf/579_Paper.pdf}
  {Conventional orthography for dialectal {A}rabic}.
\newblock In \emph{Proceedings of the Eighth International Conference on
  Language Resources and Evaluation ({LREC}'12)}, pages 711--718, Istanbul,
  Turkey. European Language Resources Association (ELRA).

\bibitem[{Hauer et~al.(2014)Hauer, Hayward, and
  Kondrak}]{hauer-etal-2014-solving}
Bradley Hauer, Ryan Hayward, and Grzegorz Kondrak. 2014.
\newblock \href {https://www.aclweb.org/anthology/C14-1218} {Solving
  substitution ciphers with combined language models}.
\newblock In \emph{Proceedings of {COLING} 2014, the 25th International
  Conference on Computational Linguistics: Technical Papers}, pages 2314--2325,
  Dublin, Ireland. Dublin City University and Association for Computational
  Linguistics.

\bibitem[{He et~al.(2020)He, Wang, Neubig, and Berg-Kirkpatrick}]{he2020a}
Junxian He, Xinyi Wang, Graham Neubig, and Taylor Berg-Kirkpatrick. 2020.
\newblock \href {https://openreview.net/forum?id=HJlA0C4tPS} {A probabilistic
  formulation of unsupervised text style transfer}.
\newblock In \emph{International Conference on Learning Representations}.

\bibitem[{Hieber and Riezler(2015)}]{hieber-riezler-2015-bag}
Felix Hieber and Stefan Riezler. 2015.
\newblock \href {https://doi.org/10.3115/v1/N15-1123} {Bag-of-words forced
  decoding for cross-lingual information retrieval}.
\newblock In \emph{Proceedings of the 2015 Conference of the North {A}merican
  Chapter of the Association for Computational Linguistics: Human Language
  Technologies}, pages 1172--1182, Denver, Colorado. Association for
  Computational Linguistics.

\bibitem[{{Hinton}(2002)}]{hinton2002poe}
G.~E. {Hinton}. 2002.
\newblock Training products of experts by minimizing contrastive divergence.
\newblock \emph{Neural Computation}, 14(8):1771--1800.

\bibitem[{Hochreiter and Schmidhuber(1997)}]{hochreiter1997long}
Sepp Hochreiter and J{\"u}rgen Schmidhuber. 1997.
\newblock Long short-term memory.
\newblock \emph{Neural computation}, 9(8):1735--1780.

\bibitem[{Holtzman et~al.(2020)Holtzman, Buys, Du, Forbes, and
  Choi}]{holtzman2020the}
Ari Holtzman, Jan Buys, Li~Du, Maxwell Forbes, and Yejin Choi. 2020.
\newblock \href {https://openreview.net/forum?id=rygGQyrFvH} {The curious case
  of neural text degeneration}.
\newblock In \emph{International Conference on Learning Representations}.

\bibitem[{Johny et~al.(2021)Johny, Wolf-Sonkin, Gutkin, and
  Roark}]{johny-etal-2021-finite}
Cibu Johny, Lawrence Wolf-Sonkin, Alexander Gutkin, and Brian Roark. 2021.
\newblock \href {https://www.aclweb.org/anthology/2021.eacl-demos.3}
  {Finite-state script normalization and processing utilities: The {N}isaba
  {B}rahmic library}.
\newblock In \emph{Proceedings of the 16th Conference of the European Chapter
  of the Association for Computational Linguistics: System Demonstrations},
  pages 14--23, Online. Association for Computational Linguistics.

\bibitem[{Knight and Graehl(1998)}]{knight-graehl-1998-machine}
Kevin Knight and Jonathan Graehl. 1998.
\newblock \href {https://www.aclweb.org/anthology/J98-4003} {Machine
  transliteration}.
\newblock \emph{Computational Linguistics}, 24(4):599--612.

\bibitem[{Knight et~al.(2006)Knight, Nair, Rathod, and
  Yamada}]{knight-etal-2006-unsupervised}
Kevin Knight, Anish Nair, Nishit Rathod, and Kenji Yamada. 2006.
\newblock \href {https://www.aclweb.org/anthology/P06-2065} {Unsupervised
  analysis for decipherment problems}.
\newblock In \emph{Proceedings of the {COLING}/{ACL} 2006 Main Conference
  Poster Sessions}, pages 499--506, Sydney, Australia. Association for
  Computational Linguistics.

\bibitem[{Koehn et~al.(2007)Koehn, Hoang, Birch, Callison-Burch, Federico,
  Bertoldi, Cowan, Shen, Moran, Zens, Dyer, Bojar, Constantin, and
  Herbst}]{koehn-etal-2007-moses}
Philipp Koehn, Hieu Hoang, Alexandra Birch, Chris Callison-Burch, Marcello
  Federico, Nicola Bertoldi, Brooke Cowan, Wade Shen, Christine Moran, Richard
  Zens, Chris Dyer, Ond{\v{r}}ej Bojar, Alexandra Constantin, and Evan Herbst.
  2007.
\newblock \href {https://www.aclweb.org/anthology/P07-2045} {{M}oses: Open
  source toolkit for statistical machine translation}.
\newblock In \emph{Proceedings of the 45th Annual Meeting of the Association
  for Computational Linguistics Companion Volume Proceedings of the Demo and
  Poster Sessions}, pages 177--180, Prague, Czech Republic. Association for
  Computational Linguistics.

\bibitem[{Lample et~al.(2018)Lample, Conneau, Denoyer, and
  Ranzato}]{lample2018unsupervised}
Guillaume Lample, Alexis Conneau, Ludovic Denoyer, and Marc'Aurelio Ranzato.
  2018.
\newblock \href {https://openreview.net/forum?id=rkYTTf-AZ} {Unsupervised
  machine translation using monolingual corpora only}.
\newblock In \emph{International Conference on Learning Representations}.

\bibitem[{Lample et~al.(2019)Lample, Subramanian, Smith, Denoyer, Ranzato, and
  Boureau}]{lample2019multipleattribute}
Guillaume Lample, Sandeep Subramanian, Eric Smith, Ludovic Denoyer,
  Marc'Aurelio Ranzato, and Y-Lan Boureau. 2019.
\newblock \href {https://openreview.net/forum?id=H1g2NhC5KQ}
  {Multiple-attribute text rewriting}.
\newblock In \emph{International Conference on Learning Representations}.

\bibitem[{Liang and Klein(2009)}]{liang-klein-2009-online}
Percy Liang and Dan Klein. 2009.
\newblock \href {https://www.aclweb.org/anthology/N09-1069} {Online {EM} for
  unsupervised models}.
\newblock In \emph{Proceedings of Human Language Technologies: The 2009 Annual
  Conference of the North {A}merican Chapter of the Association for
  Computational Linguistics}, pages 611--619, Boulder, Colorado. Association
  for Computational Linguistics.

\bibitem[{Lin et~al.(2019)Lin, Zhu, Gormley, and Eisner}]{lin-etal-2019-neural}
Chu-Cheng Lin, Hao Zhu, Matthew~R. Gormley, and Jason Eisner. 2019.
\newblock \href {https://doi.org/10.18653/v1/N19-1024} {Neural finite-state
  transducers: Beyond rational relations}.
\newblock In \emph{Proceedings of the 2019 Conference of the North {A}merican
  Chapter of the Association for Computational Linguistics: Human Language
  Technologies, Volume 1 (Long and Short Papers)}, pages 272--283, Minneapolis,
  Minnesota. Association for Computational Linguistics.

\bibitem[{Ljube{\v{s}}i{\'c} and Klubi{\v{c}}ka(2014)}]{ljubesic-klubicka-2014}
Nikola Ljube{\v{s}}i{\'c} and Filip Klubi{\v{c}}ka. 2014.
\newblock \href {https://doi.org/10.3115/v1/W14-0405} {\{bs,hr,sr\}{W}a{C} -
  web corpora of {B}osnian, {C}roatian and {S}erbian}.
\newblock In \emph{Proceedings of the 9th Web as Corpus Workshop ({W}a{C}-9)},
  pages 29--35, Gothenburg, Sweden. Association for Computational Linguistics.

\bibitem[{Makarov et~al.(2017)Makarov, Ruzsics, and
  Clematide}]{makarov-etal-2017-align}
Peter Makarov, Tatiana Ruzsics, and Simon Clematide. 2017.
\newblock \href {https://doi.org/10.18653/v1/K17-2004} {Align and copy: {UZH}
  at {SIGMORPHON} 2017 shared task for morphological reinflection}.
\newblock In \emph{Proceedings of the {C}o{NLL} {SIGMORPHON} 2017 Shared Task:
  Universal Morphological Reinflection}, pages 49--57, Vancouver. Association
  for Computational Linguistics.

\bibitem[{Mielke et~al.(2019)Mielke, Cotterell, Gorman, Roark, and
  Eisner}]{mielke-etal-2019-kind}
Sabrina~J. Mielke, Ryan Cotterell, Kyle Gorman, Brian Roark, and Jason Eisner.
  2019.
\newblock \href {https://doi.org/10.18653/v1/P19-1491} {What kind of language
  is hard to language-model?}
\newblock In \emph{Proceedings of the 57th Annual Meeting of the Association
  for Computational Linguistics}, pages 4975--4989, Florence, Italy.
  Association for Computational Linguistics.

\bibitem[{Mohri(2009)}]{mohri2009weighted}
Mehryar Mohri. 2009.
\newblock Weighted automata algorithms.
\newblock In \emph{Handbook of weighted automata}, pages 213--254. Springer.

\bibitem[{Mohri and Riley(2002)}]{mohri2002efficient}
Mehryar Mohri and Michael Riley. 2002.
\newblock An efficient algorithm for the n-best-strings problem.
\newblock In \emph{Seventh International Conference on Spoken Language
  Processing}.

\bibitem[{Papineni et~al.(2002)Papineni, Roukos, Ward, and
  Zhu}]{papineni-etal-2002-bleu-2}
Kishore Papineni, Salim Roukos, Todd Ward, and Wei-Jing Zhu. 2002.
\newblock \href {https://doi.org/10.3115/1073083.1073135} {{BLEU}: {A} method
  for automatic evaluation of machine translation}.
\newblock In \emph{Proceedings of the 40th Annual Meeting of the Association
  for Computational Linguistics}, pages 311--318, Philadelphia, Pennsylvania,
  USA. Association for Computational Linguistics.

\bibitem[{Park et~al.(2021)Park, Zhang, Haley, Steimel, Liu, and
  Schwartz}]{10.1162/tacl_a_00365}
Hyunji~Hayley Park, Katherine~J. Zhang, Coleman Haley, Kenneth Steimel, Han
  Liu, and Lane Schwartz. 2021.
\newblock \href {https://doi.org/10.1162/tacl_a_00365} {Morphology matters: {A}
  multilingual language modeling analysis}.
\newblock \emph{Transactions of the Association for Computational Linguistics},
  9:261--276.

\bibitem[{Paulsen(2014)}]{paulsen20149}
Martin Paulsen. 2014.
\newblock Translit: {C}omputer-mediated digraphia on the {R}unet.
\newblock \emph{Digital Russia: The Language, Culture and Politics of New Media
  Communication}.

\bibitem[{Pourdamghani and Knight(2017)}]{pourdamghani-knight-2017-deciphering}
Nima Pourdamghani and Kevin Knight. 2017.
\newblock \href {https://doi.org/10.18653/v1/D17-1266} {Deciphering related
  languages}.
\newblock In \emph{Proceedings of the 2017 Conference on Empirical Methods in
  Natural Language Processing}, pages 2513--2518, Copenhagen, Denmark.
  Association for Computational Linguistics.

\bibitem[{Rastogi et~al.(2016)Rastogi, Cotterell, and
  Eisner}]{rastogi-etal-2016-weighting}
Pushpendre Rastogi, Ryan Cotterell, and Jason Eisner. 2016.
\newblock \href {https://doi.org/10.18653/v1/N16-1076} {Weighting finite-state
  transductions with neural context}.
\newblock In \emph{Proceedings of the 2016 Conference of the North {A}merican
  Chapter of the Association for Computational Linguistics: Human Language
  Technologies}, pages 623--633, San Diego, California. Association for
  Computational Linguistics.

\bibitem[{Ravi and Knight(2009)}]{ravi-knight-2009-learning}
Sujith Ravi and Kevin Knight. 2009.
\newblock \href {https://www.aclweb.org/anthology/N09-1005} {Learning phoneme
  mappings for transliteration without parallel data}.
\newblock In \emph{Proceedings of Human Language Technologies: The 2009 Annual
  Conference of the North {A}merican Chapter of the Association for
  Computational Linguistics}, pages 37--45, Boulder, Colorado. Association for
  Computational Linguistics.

\bibitem[{Roark et~al.(2012)Roark, Sproat, Allauzen, Riley, Sorensen, and
  Tai}]{opengrm}
Brian Roark, Richard Sproat, Cyril Allauzen, Michael Riley, Jeffrey Sorensen,
  and Terry Tai. 2012.
\newblock \href {https://www.aclweb.org/anthology/P12-3011} {The {O}pen{G}rm
  open-source finite-state grammar software libraries}.
\newblock In \emph{Proceedings of the {ACL} 2012 System Demonstrations}, pages
  61--66, Jeju Island, Korea. Association for Computational Linguistics.

\bibitem[{Roark et~al.(2020)Roark, Wolf-Sonkin, Kirov, Mielke, Johny,
  Demirsahin, and Hall}]{roark-etal-2020-dakshina}
Brian Roark, Lawrence Wolf-Sonkin, Christo Kirov, Sabrina~J. Mielke, Cibu
  Johny, Isin Demirsahin, and Keith Hall. 2020.
\newblock \href {https://www.aclweb.org/anthology/2020.lrec-1.294} {Processing
  {S}outh {A}sian languages written in the {L}atin script: {T}he {D}akshina
  dataset}.
\newblock In \emph{Proceedings of the 12th Language Resources and Evaluation
  Conference}, pages 2413--2423, Marseille, France. European Language Resources
  Association.

\bibitem[{Ryskina et~al.(2020)Ryskina, Gormley, and
  Berg-Kirkpatrick}]{ryskina-etal-2020-phonetic}
Maria Ryskina, Matthew~R. Gormley, and Taylor Berg-Kirkpatrick. 2020.
\newblock \href {https://doi.org/10.18653/v1/2020.acl-main.737} {Phonetic and
  visual priors for decipherment of informal {R}omanization}.
\newblock In \emph{Proceedings of the 58th Annual Meeting of the Association
  for Computational Linguistics}, pages 8308--8319, Online. Association for
  Computational Linguistics.

\bibitem[{Sennrich et~al.(2016)Sennrich, Haddow, and
  Birch}]{sennrich-etal-2016-neural}
Rico Sennrich, Barry Haddow, and Alexandra Birch. 2016.
\newblock \href {https://doi.org/10.18653/v1/P16-1162} {Neural machine
  translation of rare words with subword units}.
\newblock In \emph{Proceedings of the 54th Annual Meeting of the Association
  for Computational Linguistics (Volume 1: Long Papers)}, pages 1715--1725,
  Berlin, Germany. Association for Computational Linguistics.

\bibitem[{Shavrina and Shapovalova(2017)}]{shavrina2017methodology}
Tatiana Shavrina and Olga Shapovalova. 2017.
\newblock To the methodology of corpus construction for machine learning:
  {T}aiga syntax tree corpus and parser.
\newblock In \emph{Proc. CORPORA 2017 International Conference}, pages 78--84,
  St. Petersburg.

\bibitem[{Song et~al.(2014)Song, Strassel, Lee, Walker, Wright, Garland, Fore,
  Gainor, Cabe, Thomas, Callahan, and Sawyer}]{song-etal-2014-collecting}
Zhiyi Song, Stephanie Strassel, Haejoong Lee, Kevin Walker, Jonathan Wright,
  Jennifer Garland, Dana Fore, Brian Gainor, Preston Cabe, Thomas Thomas,
  Brendan Callahan, and Ann Sawyer. 2014.
\newblock \href
  {http://www.lrec-conf.org/proceedings/lrec2014/pdf/1094_Paper.pdf}
  {Collecting natural {SMS} and chat conversations in multiple languages: The
  {BOLT} phase 2 corpus}.
\newblock In \emph{Proceedings of the Ninth International Conference on
  Language Resources and Evaluation ({LREC}'14)}, pages 1699--1704, Reykjavik,
  Iceland. European Language Resources Association (ELRA).

\bibitem[{Stahlberg and Byrne(2019)}]{stahlberg-byrne-2019-nmt}
Felix Stahlberg and Bill Byrne. 2019.
\newblock \href {https://doi.org/10.18653/v1/D19-1331} {On {NMT} search errors
  and model errors: Cat got your tongue?}
\newblock In \emph{Proceedings of the 2019 Conference on Empirical Methods in
  Natural Language Processing and the 9th International Joint Conference on
  Natural Language Processing (EMNLP-IJCNLP)}, pages 3356--3362, Hong Kong,
  China. Association for Computational Linguistics.

\bibitem[{Wu and Cotterell(2019)}]{wu-cotterell-2019-exact}
Shijie Wu and Ryan Cotterell. 2019.
\newblock \href {https://doi.org/10.18653/v1/P19-1148} {Exact hard monotonic
  attention for character-level transduction}.
\newblock In \emph{Proceedings of the 57th Annual Meeting of the Association
  for Computational Linguistics}, pages 1530--1537, Florence, Italy.
  Association for Computational Linguistics.

\bibitem[{Wu et~al.(2021)Wu, Cotterell, and Hulden}]{wu-etal-2021-applying}
Shijie Wu, Ryan Cotterell, and Mans Hulden. 2021.
\newblock \href {https://www.aclweb.org/anthology/2021.eacl-main.163} {Applying
  the transformer to character-level transduction}.
\newblock In \emph{Proceedings of the 16th Conference of the European Chapter
  of the Association for Computational Linguistics: Main Volume}, pages
  1901--1907, Online. Association for Computational Linguistics.

\bibitem[{Wu et~al.(2018)Wu, Shapiro, and Cotterell}]{wu-etal-2018-hard}
Shijie Wu, Pamela Shapiro, and Ryan Cotterell. 2018.
\newblock \href {https://doi.org/10.18653/v1/D18-1473} {Hard non-monotonic
  attention for character-level transduction}.
\newblock In \emph{Proceedings of the 2018 Conference on Empirical Methods in
  Natural Language Processing}, pages 4425--4438, Brussels, Belgium.
  Association for Computational Linguistics.

\bibitem[{Yang et~al.(2018)Yang, Hu, Dyer, Xing, and
  Berg-Kirkpatrick}]{yang2018unsupervised}
Zichao Yang, Zhiting Hu, Chris Dyer, Eric~P. Xing, and Taylor Berg-Kirkpatrick.
  2018.
\newblock \href
  {http://papers.nips.cc/paper/7959-unsupervised-text-style-transfer-using-language-models-as-discriminators}
  {Unsupervised text style transfer using language models as discriminators}.
\newblock In \emph{NeurIPS}, pages 7298--7309.

\end{thebibliography}
\clearpage

\appendix

\section{Data download links}
\label{app:download}
The romanized Russian and Arabic data and preprocessing scripts can be downloaded \href{https://github.com/ryskina/romanization-decipherment}{here}. This repository also contains the relevant portion of the Taiga dataset, which can be downloaded in full \href{https://tatianashavrina.github.io/taiga_site/downloads}{at this link}. The romanized Kannada data was downloaded from the \href{https://github.com/google-research-datasets/dakshina}{Dakshina dataset}.

The scripts to download the Serbian and Bosnian Leipzig corpora data can be found \href{https://github.com/cindyxinyiwang/deep-latent-sequence-model}{here}. The UDHR texts were collected from the corresponding pages: \href{https://unicode.org/udhr/d/udhr_srp_cyrl.txt}{Serbian}, \href{https://unicode.org/udhr/d/udhr_bos_latn.txt}{Bosnian}.

The keyboard layouts used to construct the phonetic priors are collected from the following sources: \href{http://arabic.omaralzabir.com/}{Arabic 1},  \href{https://thomasplagwitz.com/2013/01/06/imrans-phonetic-keyboard-for-arabic/}{Arabic 2}, \href{http://winrus.com/kbd_e.htm}{Russian}, \href{http://kaulonline.com/uninagari/kannada/}{Kannada},
\href{http://ascii-table.com/keyboard.php/450}{Serbian}. The Unicode confusables list used for the Russian visual prior can be found \href{https://www.unicode.org/Public/security/latest/confusables.txt}{here}.

\section{Implementation}
\label{app:implementation}

\paragraph{WFST}
We reuse the unsupervised WFST implementation of~\citet{ryskina-etal-2020-phonetic},\footnote{
\url{https://github.com/ryskina/romanization-decipherment}} which utilizes the OpenFst~\cite{openfst} and OpenGrm~\cite{opengrm} libraries. We use the default hyperparameter settings 
described by the authors (see Appendix B in the original paper).
We keep the hyperparameters unchanged for the translation experiment and set the maximum delay value to 2 for both translation directions.

\paragraph{UNMT}
We use the PyTorch UNMT implementation of~\citet{he2020a}\footnote{\url{https://github.com/cindyxinyiwang/deep-latent-sequence-model}} which incorporates improvements introduced by \citet{lample2019multipleattribute} such as the addition of a max-pooling layer. We use a single-layer LSTM~\cite{hochreiter1997long} with hidden state size 512 for both the encoder and the decoder and embedding dimension 128. For the denoising autoencoding loss, we adopt the default noise model and hyperparameters as described by~\citet{lample2018unsupervised}. The autoencoding loss is annealed over the first 3 epochs. We predict the output using greedy decoding and set the maximum output length equal to the length of the input sequence. Patience for early stopping is set to 10.

\paragraph{Model combinations}
Our joint decoding implementations rely on PyTorch and the Pynini finite-state library~\cite{gorman2016pynini}. In reranking, we rescore $n=5$ best hypotheses produced using beam search and $n$--shortest path algorithm for the UNMT and WFST respectively. Product of experts decoding is also performed with beam size 5.

\section{Metrics}
\label{app:metrics}
The character error rate (CER) and word error rate (WER) as measured as the Levenshtein distance between the hypothesis and reference divided by reference length:
$$
\mathrm{ER}(h, r) = \frac{dist(h, r)}{len(r)}
$$
with both the numerator and the denominator measured in characters and words respectively. 

We report BLEU-4 score~\cite{papineni-etal-2002-bleu-2}, measured using the Moses toolkit script.\footnote{\url{https://github.com/moses-smt/mosesdecoder/blob/master/scripts/generic/multi-bleu.perl}} For both BLEU and WER, we split sentences into words using the Moses tokenizer~\cite{koehn-etal-2007-moses}.

\clearpage

\begin{table*}[htp]
    \centering
    \small
    \setlength\tabcolsep{2pt}
    \resizebox{\textwidth}{!}{
    \begin{tabular}{p{1in}>{\raggedright\arraybackslash}p{2.5in}>{\raggedright\arraybackslash}p{2.5in}}
        \toprule
        \rowcolor{light-gray} Input & 
            kongress ne odobril biudjet dlya osuchestvleniye "bor'bi s kommunizmom" v yuzhniy amerike. &
        \\
        Ground truth & 
            \textcyr{kongress ne odobril byudzhet dlya osushchestvleniya} "\textcyr{borp1by s kommunizmom}" \textcyr{v yuzhno\u{i} amerike.} &
            kongress ne odobril bjud\v{z}et dlja osu\v{s}\v{c}estvlenija "bor'by s kommunizmom" v ju\v{z}noj amerike.
        \\
        \midrule
        \rowcolor{light-gray} WFST & 
          \textcyr{kongress ne odobril \errorbox{viu}d\gap{}et dl\errorbox{a} osu\errorbox{sch}estvleni\errorbox{y}\translitbox{e}} "\textcyr{bor}\errorbox{\#}\textcyr{b\errorbox{i} s kommunizmom}" \textcyr{v \errorbox{uuz}n\errorbox{ani} amerike.} &
          kongress ne odobril \errorbox{viu}d\gap{}et dl\errorbox{a} osu\errorbox{s\v{c}}estvleni\errorbox{y}\translitbox{e} "bor\errorbox{\#}b\errorbox{i} s kommunizmom" v \errorbox{uuz}n\errorbox{ani} amerike.
        \\
        Reranked WFST &
            \textcyr{kongress ne odobril \errorbox{vi}d\errorbox{\vphantom{0}}et d\errorbox{e}l\errorbox{a} osu\errorbox{sch}estvleni\errorbox{y}\translitbox{e}} "\textcyr{bor}\errorbox{\#}\textcyr{b\errorbox{i} s kommunizmom}" \textcyr{v \errorbox{uuz}n\errorbox{ani} amerike.} &
            kongress ne odobril \errorbox{vi}d\gap{}et d\errorbox{e}l\errorbox{a} osu\errorbox{s\v{c}}estvleni\errorbox{y}\translitbox{e} "bor\errorbox{\#}b\errorbox{i} s kommunizmom" v \errorbox{uuz}n\errorbox{ani} amerike.
        \\
        \rowcolor{light-gray} Seq2Seq &
            \textcyr{kongress ne odobril b\errorbox{y udivitelp1no \phantom{0000000000}} s kommunizmom}" \textcyr{v yuzhn\translitbox{y}\u{i} amerike.} &
            kongress ne odobril b\errorbox{y udivitel'no \phantom{00000000000}} \errorbox{\phantom{0000000}} s kommunizmom" v ju\v{z}n\translitbox{y}j amerike.
        \\
        Reranked Seq2Seq &
            \textcyr{kongress ne odobril byudzhet dlya osushchestvleni\translitbox{e}} "\textcyr{borp1by s kommunizmom}" \textcyr{v yuzhn\translitbox{y}\u{i} amerike.} &
            kongress ne odobril bjud\v{z}et dlja osu\v{s}\v{c}estvleni\translitbox{e} "bor'by s kommunizmom" v ju\v{z}n\translitbox{y}j amerike.
        \\
        \rowcolor{light-gray} Product of experts &
            \textcyr{kongress ne odobril b\errorbox{i}d\gap{}et dlya\errorbox{ a} osushchestvleni\errorbox{y}\translitbox{e}} "\textcyr{borp1by s kommunizmom}" \textcyr{v \errorbox{uuz}n\errorbox{nik} ameri\errorbox{\phantom{00}}} &
            kongress ne odobril b\errorbox{i}d\gap{}et dlja\errorbox{ a} osu\v{s}\v{c}estvleni\errorbox{y}\translitbox{e} "bor'by s kommunizmom" v \errorbox{uuz}n\errorbox{nik} ameri\errorbox{\phantom{00}}
        \\
        \bottomrule
    \end{tabular}
    }
    \caption{Different model outputs for a Russian transliteration example (left column---Cyrillic, right---scientific transliteration).  Prediction errors are shown in \errorbox{red}. Correctly transliterated segments that do not match the ground truth because of spelling standardization in annotation are in \translitbox{yellow}. \# stands for UNK.}
\end{table*}

\begin{table*}[htp]
    \centering
    \small
    \setlength\tabcolsep{2pt}
    \begin{tabular}{p{1.2in}p{2.2in}p{2.2in}}
        \toprule
        \rowcolor{light-gray} Input & 
             ana h3dyy 3lek bokra 3la 8 kda &
        \\
        Ground truth & 
            \<كده> 
            8 
            \<انا حأعدي عليك بكرة على>
            & AnA H>Edy Elyk bkrp ElY 8 kdh
        \\
        \midrule
        \rowcolor{light-gray} WFST & 
            \<كده> 
            8
            \<انا حد يي لك بكر لأ > 
            & AnA H\gap{}d\errorbox{\phantom{0}}y\translitbox{y} \gap{}l\gap{}k bkr\gap{}  l\errorbox{>} 8 kdh
        \\
        Reranked WFST &
            \<كده>
            8
            \<انا حد يي لك بكر لأ>            
            & AnA H\gap{}d\errorbox{\phantom{0}}y\translitbox{y} \gap{}l\gap{}k bkr\gap{}  l\errorbox{>} 8 kdh
        \\
        \rowcolor{light-gray} Seq2Seq &
            \<كده>
            1
            \<انا بأدي أخلك حر أول>
            & AnA \errorbox{b}>dy \errorbox{>x}l\gap{}k 
            \errorbox{H}r\gap{} \errorbox{>w}l\gap{} \errorbox{1} kdh
        \\
        Reranked Seq2Seq &
            \<كده>
            1
            \<انا بأدي أخلك حر أول>  
            & AnA \errorbox{b}>dy \errorbox{>x}l\gap{}k 
            \errorbox{H}r\gap{} \errorbox{>w}l\gap{} \errorbox{1} kdh
        \\
        \rowcolor{light-gray} Product of experts & 
            \<كده>
            8
            \<انا دي لك ب كرا ألا>
            & AnA \gap{}dy \gap{}l\gap{}k b\errorbox{\phantom{0}}kr\errorbox{A} \errorbox{> }l\errorbox{A} 8 kdh
        \\
        \bottomrule
    \end{tabular}
    \caption{Different model outputs for an Arabizi transliteration example (left column---Arabic, right---Buckwalter transliteration).  Prediction errors are highlighted in \errorbox{red} in the romanized versions. Correctly transliterated segments that do not match the ground truth because of spelling standardization during annotation are highlighted in \translitbox{yellow}.}
\end{table*}

\begin{table*}[htp]
    \centering
    \small
    \setlength\tabcolsep{2pt}
    \resizebox{\textwidth}{!}{
    \begin{tabular}{p{1.1in}p{3.3in}>{\raggedright\arraybackslash}p{2in}}
        \toprule
        \rowcolor{light-gray} Input & 
              kshullaka baalina avala horaatavannu adu vivarisuttade.
            &
        \\
        Ground truth & 
           \raisebox{-.3\height}{\includegraphics[height=0.5cm]{ 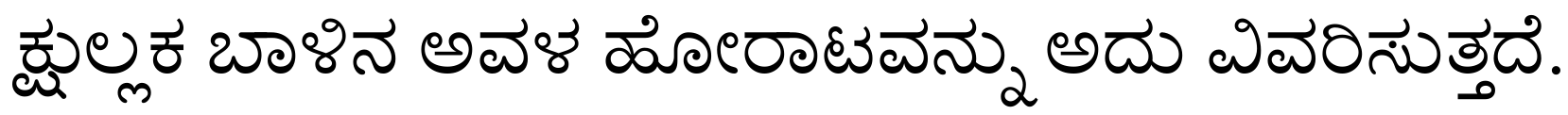}}
            & k\d{s}ullaka b\={a}\d{l}ina ava\d{l}a h\={o}r\={a}\d{t}avannu adu vivarisuttade.
        \\
        \midrule
        \rowcolor{light-gray} WFST & 
            \raisebox{-.3\height}{\includegraphics[height=0.49cm]{ 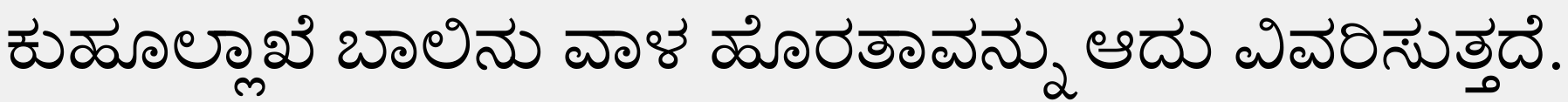}}
            & k\gap{}u\errorbox{h\={u}}ll\errorbox{\={a}}k\errorbox{he} b\={a}\errorbox{l}in\errorbox{u} 
            \gap{}v\errorbox{\={a}}\d{l}a h\errorbox{o}r\errorbox{at\={a}}vannu 
            \errorbox{\={a}}du 
            vivarisuttade.
        \\
        Reranked WFST &
            \raisebox{-.3\height}{\includegraphics[height=0.5cm]{ 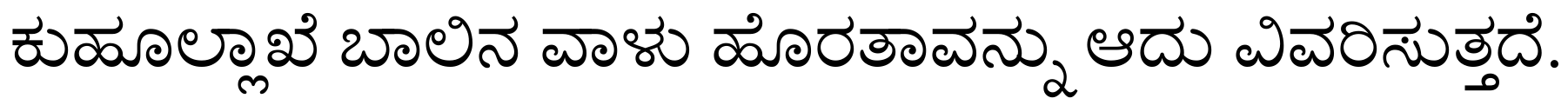}}
            & k\gap{}u\errorbox{h\={u}}ll\errorbox{\={a}}k\errorbox{he} b\={a}\errorbox{l}ina 
            \gap{}v\errorbox{\={a}}\d{l}\errorbox{u} 
            h\errorbox{o}r\errorbox{at\={a}}vannu 
            \errorbox{\={a}}du 
            vivarisuttade.
        \\
        \rowcolor{light-gray} Seq2Seq &
            \raisebox{-.3\height}{\includegraphics[height=0.5cm]{ 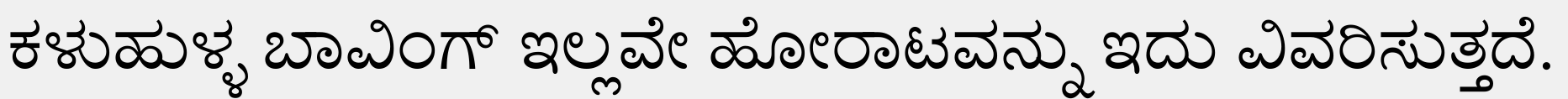}}
            & k\errorbox{a\d{l}}u\errorbox{hu\d{l}\d{l}}a
            b\={a}\errorbox{v}i\errorbox{\.{m}g}
            \errorbox{ill}av\errorbox{\={e}}
            h\={o}r\={a}\d{t}avannu
            \errorbox{i}du
            vivarisuttade.
        \\
        Reranked Seq2Seq &
            \raisebox{-.3\height}{\includegraphics[height=0.47cm]{ 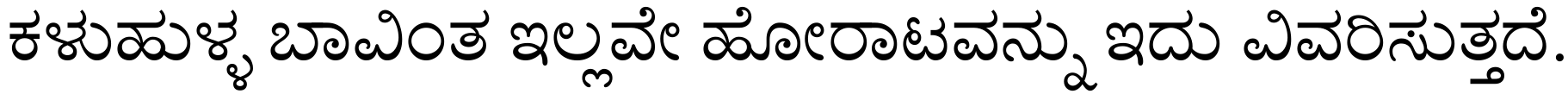}}
            & k\errorbox{a\d{l}}u\errorbox{hu\d{l}\d{l}}a
            b\={a}\errorbox{v}i\errorbox{\.{m}t}a 
            \errorbox{ill}av\errorbox{\={e}}
            h\={o}r\={a}\d{t}avannu
            \errorbox{i}du
            vivarisuttade.
        \\
        \rowcolor{light-gray} Product of experts & 
            \raisebox{-.3\height}{\includegraphics[height=0.5cm]{ 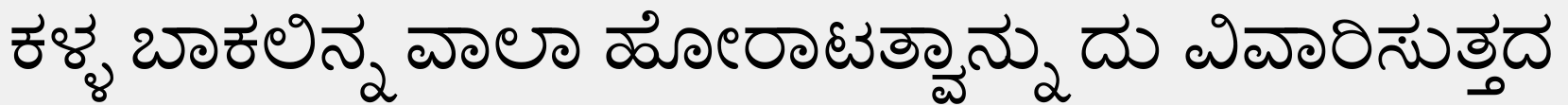}}
            & k\gap{}a\errorbox{\d{l}\d{l}}a 
            b\={a}\errorbox{kal}in\errorbox{n}a
            \gap{}v\errorbox{\={a}l\={a}}
            h\={o}r\={a}\d{t}a\errorbox{t}v\errorbox{\={a}}nnu
            \gap{}du
            viv\errorbox{\={a}}risuttad\errorbox{a\phantom{0}}
        \\
        \bottomrule
    \end{tabular}
    }
    \caption{Different model outputs for a Kannada transliteration example (left column---Kannada, right---ISO 15919 transliterations). The ISO romanization is generated using the Nisaba library~\cite{johny-etal-2021-finite}.  Prediction errors are highlighted in \errorbox{red} in the romanized versions.}
\end{table*}

\end{document}